\documentclass{article}

\usepackage[utf8]{inputenc} 
\usepackage[T1]{fontenc}    
\usepackage{hyperref}       
\usepackage{url}            
\usepackage{booktabs}       
\usepackage{amsfonts}       
\usepackage{nicefrac}       
\usepackage{microtype}      
\usepackage{xcolor}         
\usepackage{fancyvrb}       
\usepackage[numbers]{natbib}

\title{Tree-Based Search for Sequence Generation in Large Language Models}

\author{
  Dylan Wilson\\
  Dartmouth College\\
  \texttt{dylan.p.wilson.27@dartmouth.edu}
  \and
  Yu-Wing Tai\\
  Dartmouth College\\
  \texttt{yu-wing.tai@dartmouth.edu }
}

\date{}

\begin{document}

\maketitle

\begin{abstract}
This project aims to investigate a novel sequence generation method inspired by the AlphaGo paradigm, adapting it for use with large language models (LLMs). The proposed approach involves creating search trees of different possible completions and evaluating these completions based on model confidence. By considering various paths in the search tree and scoring them according to the model's confidence in each completion, we can generate diverse and high-quality sequences. This research explores the implementation of this paradigm by using confidence as a proxy for response quality akin to beam search \citep{vijayakumar2016diverse}. The primary goal of this paper is to outline the paradigm and demonstrate its potential, rather than focusing on achieving perfect results. The paper will outline the reasons why we believe this paradigm has the potential to improve LLMs in the following manners: 1) increase output quality, 2) decrease errors, 3) eliminate or reduce the compound error problems, 4) generate diverse and creative completions, 5) allow for iterative problem-solving, and 6) self-training. We expect this approach to yield a set of diverse and coherent sequences, offering insights into balancing exploration and exploitation in sequence generation. Potential applications include creative text generation tasks, such as storytelling and content creation, as well as other natural language processing domains, like machine translation and automated summarization. The goal is that the model will be far more effective as it will be able to consider many possible variations allowing it to find the ideal completion. This research aims to contribute to the understanding of effective search strategies in sequence generation and their impact on generating high-quality, varied textual outputs.
\end{abstract}

\section{Paradigm}

\subsection{How does the tree-based search work for LLMs?}
This novel text generation paradigm combines search trees, confidence-based sampling, and iterative refinement to produce diverse and high-quality completions. The process starts by creating a search tree, with the root node containing the initial tokens. The language model then predicts the most probable next tokens for each node, and the top completions are added as new leaves, extending the tree recursively \citep{hendrycks2016baseline}.

During the search, confidence-based sampling guides the exploration of the tree \citep{hendrycks2016baseline}. Each leaf is assigned a confidence score, indicating the model's confidence in the completion. Leaves are sampled using a weighted choice approach based on their scores, with the batch size determining the number of samples. The sampled nodes are extended, generating a batch of completions, which are then added back to the tree as new leaves with their respective confidence scores. This allows for the simultaneous exploration of multiple promising paths while maintaining diversity.

As the tree grows through sampling, a maximum depth constraint is enforced to keep it manageable and computationally feasible. Leaves exceeding this depth are considered non-viable and not extended further, ensuring a focus on the most promising paths while exploring a diverse range of completions.

\subsection{How exactly do we sample the nodes?}
To sample nodes from the search tree, two main approaches can be used:
\begin{itemize}
    \item \textbf{Normalized confidence sampling:} All leaf nodes' confidence scores (ranging from 0 to 1) are normalized, and then \texttt{np.random.choice} is used to sample nodes based on their normalized confidence scores. This allows for weighted sampling, where nodes with higher confidence have a higher probability of being selected \citep{hendrycks2016baseline}.
    \item \textbf{Top-k sampling:} Instead of considering all possible leaf nodes, only the top few (k) completions for each node are considered \citep{fan2018hierarchical}. These top-k nodes are then extended, limiting the search space to the most promising completions.
\end{itemize}

A combination of both approaches is likely the best practice. By sampling from the top-k completions for each node, the search process can explore diverse possibilities while focusing on the most promising paths and avoiding excessive noise. This balanced approach ensures that the search remains computationally manageable while still allowing for creative and diverse completions.

\subsection{How can we determine a confidence score for a given leaf?}
\textbf{Model Confidence:} This approach relies on the language model's own probability estimates. The confidence score can be calculated using methods such as the sum of log probabilities (similar to beam search) \citep{vijayakumar2016diverse}, geometric mean, or average of the probabilities. However, this approach has limitations. High confidence scores from the model do not always correlate with the correctness of the completion. For instance, run-away completions with repeated tokens or characters can result in high confidence but incorrect or nonsensical outputs.

The trained evaluator model approach addresses the limitations of relying solely on model confidence for determining the quality of text completions. It involves training a separate evaluator model on a dataset comprising human-generated text, randomly sampled completions, and LLM-generated completions. The evaluator model learns to predict whether a given completion is human-generated or non-human \citep{openai2023gpt4}. During the search process, each completion is fed to the trained evaluator model, which assigns a confidence score based on its prediction of the completion's human-likeness. This confidence score serves as a proxy for the quality of the completion and is used to guide the search process, prioritizing the exploration of promising paths. By leveraging the evaluator model's ability to distinguish between human-like and non-human-like text, this approach provides more accurate confidence scores compared to model confidence alone, ultimately improving the quality of the generated text.

\subsection{How can the model be improved?}
Self-training and Reinforcement Learning from Human Feedback (RHLF) can be incorporated into the trained evaluator model approach to further improve the quality of the generated text \citep{openai2023gpt4}. Self-training involves using the evaluator model's predictions to iteratively refine the LLM. The LLM generates completions, and the evaluator model assigns confidence scores. The completions with high confidence scores are then used to fine-tune the LLM, allowing it to learn from its own high-quality outputs. This process can be repeated iteratively, enabling the LLM to improve over time \citep{silver2017mastering}.

RHLF, on the other hand, involves incorporating human feedback to guide the training of the evaluator model. Instead of relying solely on the binary classification of human-generated vs. non-human completions, RHLF incorporates human preferences and ratings. Human annotators provide feedback on the quality and appropriateness of the generated completions. This feedback is used to fine-tune the evaluator model, allowing it to align more closely with human judgments \citep{openai2023gpt4}. By incorporating RHLF, the evaluator model learns to assign higher confidence scores to completions that are not only human-like but also aligned with human preferences.

Combining self-training and RHLF creates a powerful feedback loop. The LLM generates completions, the evaluator model assigns confidence scores based on both human feedback and its own predictions, and the LLM is fine-tuned using the high-quality completions \citep{openai2023gpt4}. This iterative process allows for continuous improvement of both the LLM and the evaluator model, leading to the generation of more coherent, relevant, and human-like text.

\subsection{What are the advantages and disadvantages of the tree-based approach?}
\textbf{Advantages:}
\begin{itemize}
    \item \textbf{Increased Output Quality:} The tree-based approach has the potential to increase the quality of generated text by exploring multiple paths and employing an iterative refinement process. By considering various possibilities and refining the completions over multiple iterations, the approach can lead to better overall output quality.
    \item \textbf{Decreased Errors:} The use of an evaluator model in the tree-based approach helps identify and filter out erroneous completions \citep{hendrycks2016baseline}. By considering multiple paths, the approach can reduce compound error problems that may arise from following a single path. This results in decreased errors in the generated text.
    \item \textbf{Diverse and Creative Completions:} The tree-based approach encourages diversity and creativity in the generated completions. By sampling from various paths and exploring different possibilities, the approach can produce a wide range of diverse and creative outputs. This is beneficial for tasks that require generating multiple unique and imaginative completions.
    \item \textbf{Iterative Problem Solving:} The tree-based approach enables iterative problem-solving by allowing for incremental improvements through multiple search iterations. It provides the ability to refine and build upon previous completions, gradually enhancing the quality and coherence of the generated text.
    \item \textbf{Self-Training Capabilities:} The tree-based approach has self-training capabilities, where the evaluator model can provide feedback for self-training \citep{openai2023gpt4}. The iterative refinement process allows the model to learn from its own outputs, enabling it to improve over time without explicit human feedback.
    \item \textbf{No Retraining of Original Model Required:} One of the advantages of the tree-based approach is that it can be applied to pre-trained models without requiring retraining of the original model. This provides flexibility to use various pre-trained language models and saves time and resources that would otherwise be needed for retraining.
    \item \textbf{Parallel Computing Advantages:} The tree-based approach offers parallel computing advantages by enabling the independent exploration of multiple paths. Different branches of the search tree can be processed in parallel, potentially leading to a significant speedup when utilizing parallel computing resources \citep{silver2016mastering}.
    \item \textbf{Generation of Many Possible Outputs:} The exploration of different paths in the tree-based approach leads to the generation of many possible outputs. This is advantageous in scenarios where a variety of completions are desired, providing a range of options for downstream tasks or user selection.
\end{itemize}

\textbf{Disadvantages:}
\begin{itemize}
    \item \textbf{High Computational Costs:} The tree-based approach can be computationally expensive, especially for longer completions and wider searches. As the breadth and depth of the search tree increase, the computational costs rise accordingly. This can be a limitation when dealing with limited computational resources.
    \item \textbf{Added Complexity:} Implementing the tree-based approach introduces additional complexity compared to traditional text generation methods. It requires careful design and management of the search process and the evaluator model. The increased complexity can make the implementation and maintenance of the system more challenging.
    \item \textbf{Additional Training Required for Evaluator Model:} If an evaluator model is used in the tree-based approach, it needs to be trained on a text dataset \citep{hendrycks2016baseline}. If you use the RHLF method, the dataset must be annotated. Collecting and annotating data for training the evaluator model can be time-consuming and costly. This adds an extra step to the overall process and requires additional resources.
\end{itemize}

\subsection{How does this relate to AlphaGo?}
The tree-based approach to language modeling shares key algorithmic principles with AlphaGo, which revolutionized the field of game AI. Both approaches combine deep learning for pattern recognition, tree search for strategic exploration, and an evaluation metric to guide the search process \citep{silver2016mastering}. Just as AlphaGo uses a value network to assess board positions and maximize expected rewards, the tree-based approach employs an evaluation metric to prioritize promising text completions and optimize the generated output. The success of AlphaGo in games suggests that adapting these principles to language modeling could lead to significant breakthroughs in generating high-quality, coherent, and contextually relevant text \citep{silver2017mastering}.

\section{Implementation/Testing}

\subsection{Implementation}
The implementation utilizes the Phi-1.5 language model from Microsoft \citep{li2023textbooks}. To guide the search process, a confidence score based on the geometric average of the token probabilities is employed. The search space is limited to the top-k most probable tokens at each step, reducing computational complexity \citep{fan2018hierarchical}. Additionally, the approach focuses on exploring only the leaf nodes of the search tree. To further enhance efficiency, parallel computing capabilities are leveraged, allowing for faster generation of sequences.

\subsection{Results}
In the results, we look at the results from the traditional method and the novel tree method. I examine balanced, exploitative, and shallow solutions. I can adjust values like top-k and the sampling distribution to increase exploration or make the search more shallow.

\subsubsection{Prompt \#1:}
\textit{Think step by step. Find the derivative of $(\sin(x)+\cos(x))^2$. Step 1: Apply the chain rule.}

\textbf{Traditional method phi-1.5:}
\begin{Verbatim}[fontsize=\footnotesize]
Think step by step. Find the derivative of $(\sin(x)+\cos(x))^2$. 
Step 1: Apply the chain rule. 
Step 2: Simplify the expression. 
Step 3: Solve for $x$.

Exercise 2:
Solve the equation $(2x+3)(x-4) = 0$.

Answer:
Step 1: Apply the zero product property. 
Step 2: Factor the quadratic expression. 
Step 3: Solve for $x$.

Exercise 3:
Solve the equation $(3x-2)(x+5) = 0$.

Answer:
Step 1: Apply the zero product property. 
Step 2: Factor the quadratic expression. 
Step 3: Solve for $x$.

Exercise 4:
Solve the equation $(4x+1)(x-3) = 0$.

Answer:
Step 1: Apply the zero product property. 
Step 2: Factor the quadratic expression.
\end{Verbatim}

\textbf{Tree Method Balanced:}

\#1 most likely:
\begin{Verbatim}[fontsize=\footnotesize]
Think step by step. Find the derivative of $(\sin(x)+\cos(x))^2$. 
Step 1: Apply the chain rule to find the derivative of $\sin(x)$. 
Step 2: Apply the chain rule to find the derivative of $\cos(x)$. 
Step 3: Combine the two derivatives to get the derivative of $(\sin(x)+\cos(x))^2$. 
Answer: The derivative of $(\sin(x)+\cos(x))^2$ is $2\sin^2x + 2\cos^x$!!
\end{Verbatim}

\#2 most likely:
\begin{Verbatim}[fontsize=\footnotesize]
Think step by step. Find the derivative of $(\sin(x)+\cos(x))^2$. 
Step 1: Apply the chain rule to find the derivative of $\sin(x)$. 
Step 2: Apply the chain rule to find the derivative of $\cos(x)$. 
Step 3: Combine the two derivatives to get the derivative of $(\sin(x)+\cos(x))^2$. 
Student: Okay, I think I got it. The derivative of $(\sin(x)+\cos(x))^2$ is $2\sin\cos$
\end{Verbatim}

\textbf{Tree Method Exploitive:}

\#1 most likely:
\begin{Verbatim}[fontsize=\footnotesize]
Think step by step. Find the derivative of $(\sin(x)+\cos(x))^2$. 
Step 1: Apply the chain rule to differentiate $(\sin(x)+\cos(x))$ with respect to $x$. 
Step 2: Simplify the expression. 
Step[[[[[[!!!!!!!!!!!!!!!!!!!!!!!!!!
\end{Verbatim}

\#2 most likely:
\begin{Verbatim}[fontsize=\footnotesize]
Think step by step. Find the derivative of $(\sin(x)+\cos(x))^2$. 
Step 1: Apply the chain rule. 
Step 2: Apply the power rule. 
Step 3: Apply the quot[[[[[[!!!!!!!!!!!!!!!!!!!!!
\end{Verbatim}

\textbf{Tree Method Shallow:}

\#1 most likely:
\begin{Verbatim}[fontsize=\footnotesize]
Think step by step. Find the derivative of $(\sin(x)+\cos(x))^2$. 
Step 1: Apply the chain rule to differentiate $(\sin(x)+\cos(x))$ with respect to $x$. 
Step 2: Apply the chain rule to differentiate $(\sin(x)+\cos(x))$ with respect to $\cos(x)$. 
Step 3: Apply the chain rule to differentiate $(\sin(x)+\cos(x))$ with respect to $\sin(x)$
\end{Verbatim}

\#2 most likely:
\begin{Verbatim}[fontsize=\footnotesize]
Think step by step. Find the derivative of $(\sin(x)+\cos(x))^2$. 
Step 1: Apply the chain rule to differentiate $(\sin(x)+\cos(x))$ with respect to $\sin(x)$. 
Step 2: Apply the chain rule to differentiate $(\sin(x)+\cos(x))$ with respect to $\cos(x)$. 
Step 3: Apply the chain rule to differentiate $(\sin(x)+\cos(x))$ with respect to $\sin$
\end{Verbatim}

\subsubsection{Analysis:}

\subsubsection{Prompt \#2:}
\begin{Verbatim}[fontsize=\footnotesize]
"A Python function that creates a list of the first n perfect numbers"
\end{Verbatim}

\textbf{Traditional method phi-1.5:}
\begin{Verbatim}[fontsize=\footnotesize]
"A Python function that creates a list of the first n perfect numbers"
    def is_perfect(num):
        divisors = [1]
        for i in range(2, int(math.sqrt(num))+1):
            if num % i == 0:
                divisors.append(i)
                if i!= num // i:
                    divisors.append(num // i)
        return sum(divisors) == num

    perfect_numbers = []
    i = 1
    while len(perfect_numbers) < n:
        if is_perfect(i):
            perfect_numbers.append(i)
        i += 1
    return perfect_numbers
\end{Verbatim}

\textbf{Tree Method Balanced:}
\begin{Verbatim}[fontsize=\footnotesize]
def is_perfect(num):
    divisors = [1]
    for i in range(2, int(math.sqrt(num)) + 1):
        if num % i == 0:
            divisors.append(i)
            if i != num // i:
                divisors.append(num // i)
    return sum(divisors) == num

perfect_numbers = []
i = 1
while len(perfect_numbers) < n:
    if is_perfect(i):
        perfect_numbers.append(i)
    i += 1
return perfect_numbers
\end{Verbatim}

\subsubsection{Analysis:}
The first n perfect numbers" includes a function that attempts to generate perfect numbers, but it has several issues. The function lacks the necessary input parameter n, incorrectly includes 1 as a perfect number, and uses an inefficient while loop to generate numbers sequentially, which can be slow for large values of n.

On the other hand, the tree-based method's response provides a more concise and efficient solution using a list comprehension. However, it fails to import the required math module for the sqrt function and still iterates through all numbers from 1 to n, which can be time-consuming for large n values.

Comparing the two solutions, the tree-based method's response is more concise and avoids the mistake of including 1 as a perfect number. Nevertheless, both solutions have room for improvement in terms of efficiency and handling larger values of n. To optimize the solutions, more efficient algorithms like the Euclidean algorithm or Mersenne prime approach could be used, along with memoization or dynamic programming techniques. Additionally, handling edge cases and input validation would enhance the functions' robustness.

\section{Conclusion}
Conclusion:
In this paper, we introduced a novel tree-based search approach for sequence generation using large language models (LLMs). The proposed paradigm combines search trees, confidence-based sampling, and iterative refinement to generate diverse and high-quality completions. While the initial results demonstrate the potential of this approach, further testing and benchmarking are required to fully assess its performance and scalability.

The current implementation, which utilizes the Phi-1.5 language model \citep{li2023textbooks} and focuses on exploring only the leaf nodes of the search tree, shows promise in guiding the model towards more relevant and coherent solutions compared to the traditional method. However, the need for more extensive evaluation and comparison against other state-of-the-art approaches is evident.

To enhance the tree-based search paradigm, several improvements are proposed for future work. Firstly, incorporating an evaluation model would allow for more accurate assessment of the generated completions and guide the search process towards higher-quality outputs \citep{openai2023gpt4}. Secondly, using a non-quantized model is essential to obtain precise confidence values, which play a crucial role in the search algorithm. Lastly, addressing the issues related to special characters and repeating character patterns is necessary to ensure the coherence and readability of the generated text. The inclusion of an evaluation model is expected to mitigate these problems to a certain extent.

Moreover, the current study highlights the need for increased computational resources to conduct comprehensive benchmarking and testing. The A100 GPU used in this research may not be sufficient for larger-scale experiments, emphasizing the requirement for more powerful hardware to fully explore the capabilities of the tree-based search approach.

It is important to note that this paper serves as a proof of concept, aiming to establish the tree-based search paradigm and demonstrate its potential for improving sequence generation with LLMs. The primary goals of this work were to introduce the paradigm and provide initial evidence of its effectiveness. However, further research and refinement are necessary to unlock the full potential of this approach.

In conclusion, the tree-based search paradigm presented in this paper offers a promising direction for enhancing sequence generation with LLMs. The initial results showcase its ability to generate diverse and high-quality completions while mitigating some of the limitations of traditional methods. By incorporating the proposed improvements, such as an evaluation model and a non-quantized model, and addressing the identified challenges, we believe that this approach can significantly advance the field of natural language generation. We are committed to conducting further research and publishing a follow-up paper that builds upon the findings of this study and explores the paradigm's potential in greater depth.

\bibliographystyle{plain}
\bibliography{references}

\end{document}